\documentclass[10pt,twocolumn,letterpaper]{article}

\usepackage[arXiv]{cvpr}      

\usepackage{microtype}

\makeatletter
\renewcommand{\LenToUnit}[1]{\strip@pt\dimexpr#1\relax}
\makeatother

\makeatletter
\AtBeginDocument{%
  \def\NAT@cmprs{\z@}%
  \def\NAT@sep{,}%
  \let\NAT@space\@empty
}
\makeatother

\usepackage{booktabs}
\usepackage{multirow}
\usepackage{array}
\usepackage{tabularx}
\usepackage{graphicx}
\usepackage{xcolor}
\usepackage{colortbl}
\usepackage{amsmath} 

\definecolor{rankfirst}{RGB}{255,179,179}
\definecolor{ranksecond}{RGB}{255,204,153}
\definecolor{rankthird}{RGB}{255,255,153}

\providecommand{\best}[1]{\cellcolor{rankfirst}#1}
\providecommand{\secondbest}[1]{\cellcolor{ranksecond}#1}
\providecommand{\thirdbest}[1]{\cellcolor{rankthird}#1}

\definecolor{cvprblue}{rgb}{0.21,0.49,0.74}
\usepackage[pagebackref,breaklinks,colorlinks,allcolors=cvprblue]{hyperref}

\def\confName{CVPR}
\def\confYear{2027}

\title{RAPID: Reconstruction Accelerated by Polarization-Informed Depth for 2D Gaussian Splatting}

\author{Daisong Ouyang$^{1}$, Linlin Xia$^{1}$, Liangkun Yan$^{1}$, Song Gu$^{1}$,\\
Shedong Zhao$^{1}$, Xin Dai$^{1}$, Yi Zheng$^{1}$\\
$^{1}$Northeast Electric Power University
}

\begin{document}
\twocolumn[{
  \renewcommand\twocolumn[1][]{#1}
  \maketitle

  \noindent
  \begin{minipage}{\textwidth}
    \centering
    \includegraphics[width=\linewidth]{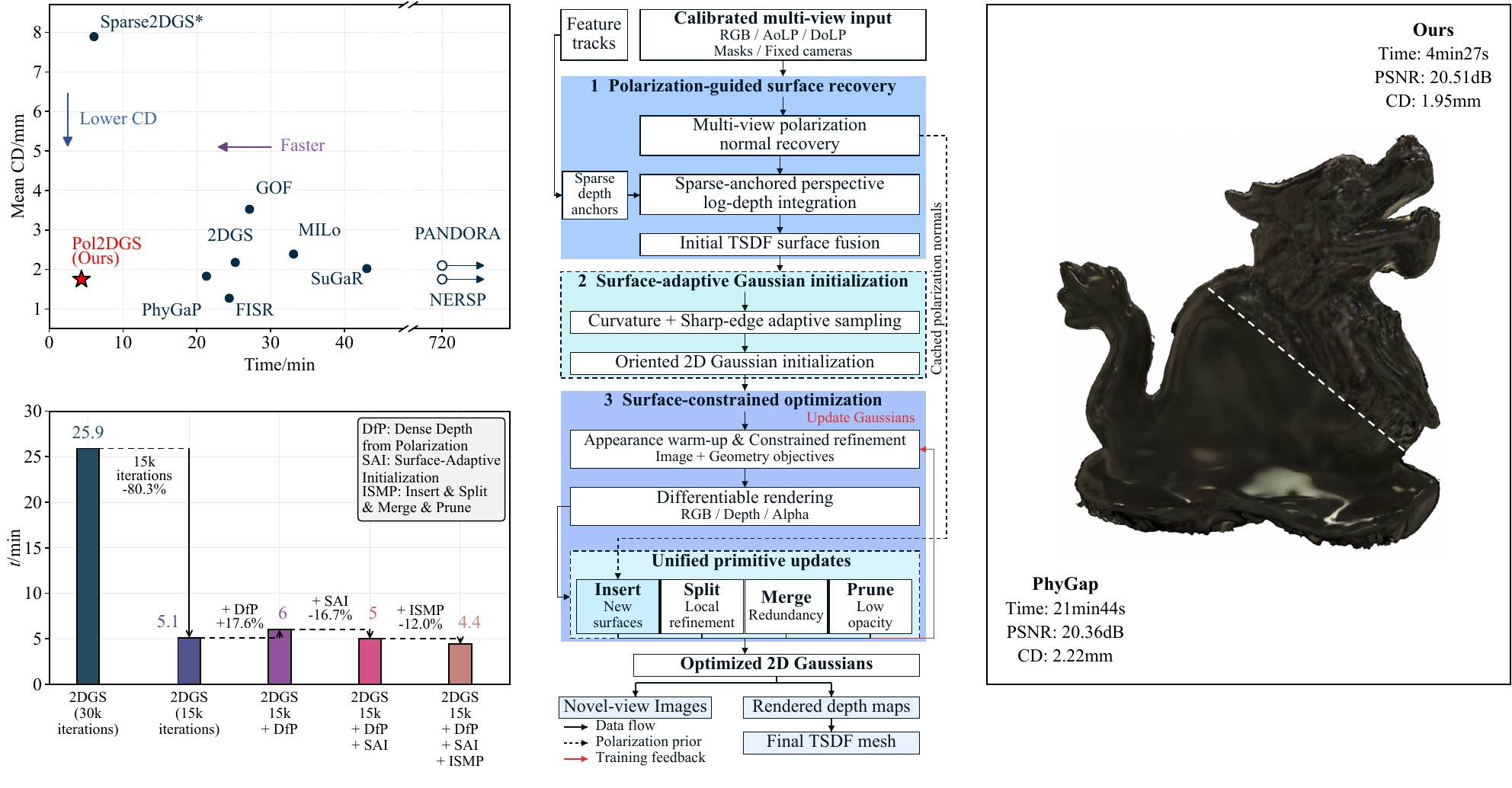}
    \captionof{figure}{\textbf{We propose Pol2DGS,} the two panels on the left illustrate how our method balances geometric accuracy and training time. The center shows the information flow of the proposed method. The right compares our novel view synthesis result with that of PhyGaP~\cite{phygap2026}, a state-of-the-art reconstruction method based on Gaussian splatting.}
    \label{fig:wide}
  \end{minipage}
  \par\vspace{1em}
}]

\begin{abstract}
Efficient and accurate 3D surface reconstruction from multi-view images remains a fundamental challenge in Gaussian splatting. Many existing methods rely on sparse geometric initialization and iterative densification, resulting in substantial optimization overhead and difficulties in balancing rendering quality and geometric accuracy, particularly in reflective and weakly textured regions. To address these limitations, we propose Pol2DGS, a geometry-first framework that uses multiview polarization cues to recover dense surface priors. We integrate surface normals into depth maps using sparse geometric anchors. These depth maps are then fused across views. Curvature-adaptive sampling and oriented Gaussian initialization then support efficient optimization. Experiments on the PISR dataset demonstrate an approximately \(5\times\) speedup over PhyGaP, which is a state-of-the-art Gaussian-based reconstruction method. Pol2DGS reduces the average optimization time from approximately 21 minutes (PhyGaP) to 4 minutes while maintaining competitive novel-view synthesis quality and mesh reconstruction accuracy. These results demonstrate the effectiveness of Pol2DGS in accelerating Gaussian optimization from polarimetric multi-view inputs.
\end{abstract}
\section{Introduction}
\label{sec:intro}

Recovering accurate 3D surfaces from multiview images~\cite{pmvs2010,mvsnet2018,volsdf2021,schoenberger2016mvs} and reproducing object appearance from unseen viewpoints are fundamental problems in 3D vision. 2D Gaussian splatting (2DGS)~\cite{huang2024_2dgs} represents local surfaces using oriented disks distributed in 3D space and efficiently synthesizes novel views through rasterization. However, rendering speed after model construction and the time required to build the model are distinct measures of efficiency~\cite{quicksplat2025,minisplatting2024}. Starting from a sparse point cloud still requires repeated adjustment of disk positions, orientations, and scales, along with densification to establish surface coverage~\cite{kerbl2023gaussians,huang2024_2dgs}. In reflective and weakly textured regions, color variations across views provide limited reliable geometric guidance for these adjustments~\cite{cui2017polarization,pisr2024}. Reducing scene optimization time while preserving novel view synthesis quality and mesh reconstruction accuracy therefore remains a challenge in Gaussian surface reconstruction.

Polarization observations provide information about surface orientation, supporting geometric estimation beyond color observations~\cite{cui2017polarization,pandora2022}. In neural implicit reconstruction, PANDORA~\cite{pandora2022} uses polarimetric inverse rendering to jointly recover object shape and separate diffuse and specular reflection components. NeRSP~\cite{nersp2024} combines Stokes vector rendering errors with multiview azimuth consistency to constrain the shape of reflective surfaces from sparse views. PISR~\cite{pisr2024} incorporates per-pixel perspective viewing rays into polarization constraints to directly constrain implicit surface normals. In explicit Gaussian reconstruction, Sparse2DGS~\cite{sparse2dgs2025} generates a dense initial point cloud from depth predicted by multiview stereo matching, then selectively updates primitives based on rendered geometry. PhyGaP~\cite{phygap2026} uses polarimetric deferred rendering to estimate geometry and material properties, supporting reflection decomposition and relighting. PolarGS~\cite{polargs2026} employs polarization information for depth completion and back-projects the completed depth to generate new Gaussians.

Inspired by these methods, we focus on geometric initialization. We investigate how polarization observations can determine the initial positions, orientations, and spatial extents of 2D Gaussians before training to reduce subsequent optimization time. Our research faces two core challenges:

The first challenge is recovering surface positions. Since polarization-derived normals specify tangent-plane orientations but not positions along viewing rays. Under perspective projection, normals constrain log-depth gradients, so integration recovers only relative depth~\cite{normalintegration2018}; we use sparse geometric anchors to fix its scale. Cross-view scale inconsistencies can place back-projected points on different depth layers despite accurate normals~\cite{normalintegration2018,cao2021normalintegration}, introducing position errors into Gaussian initialization and subsequent training. Depth anchoring and multiview surface alignment are therefore both necessary. A further challenge is the mismatch between depth sampling density and required primitive density~\cite{luo2018density,minisplatting2024}. Assigning one Gaussian to each depth sample creates redundant coverage in low-curvature regions, while uniformly coarser sampling risks losing detail in curved regions. A single scale is also insufficient when curvature differs between the two principal directions. Initializing representative primitives from dense observations using curvature adaptive spacing and scales emerges as a viable option. During training, merging redundant primitives and subdividing inadequately represented regions constitutes a principled strategy~\cite{gauhuman2024,bai2025memgs,rotabulo2024densification}, so that the primitive count reflects local surface requirements.

In light of these considerations, we propose Pol2DGS, a method that uses polarization-derived surface priors to accelerate 2D Gaussian optimization. Given calibrated multiview color images, polarization observations, and foreground masks, we first estimate polarization-derived normals and their confidence, and then we convert these normals into perspective log-depth gradients and recover depth for each view through least-squares integration with sparse geometric anchors. After multiview consistency checks, we fuse the depth maps using a truncated signed distance function (TSDF)~\cite{curless1996volumetric} and extract an initial surface. We then fit a quadratic surface in local tangent-plane coordinates. The two principal curvatures determine the sampling spacing along their respective principal directions. Sampled positions define Gaussian centers, while surface normals and principal directions determine disk orientations. Local sampling spacing initializes the two tangential scales. Training thus begins with a set of oriented disks that already provides surface coverage. The primitive distribution is then refined according to observations and local geometry.

Our main contributions are as follows:
\begin{itemize}
\item We integrate polarimetric normals with sparse geometric anchors, then use anisotropic, curvature-adaptive surface sampling to initialize Gaussian positions, orientations, and scales without depth sensors.
\item We replace the native 2DGS~\cite{huang2024_2dgs} primitive update strategy with insertion guided by surface geometry, deterministic splitting along major axes, merging subject to coverage constraints, and pruning based on opacity.
\item Our Pol2DGS reduces average measured runtime from roughly 21 to 4 minutes (5× speedup) while maintaining competitive novel-view synthesis quality and mesh reconstruction accuracy.
\end{itemize}

\section{Related Work}
\label{sec:related_work}

\begin{figure*}[t]
  \centering
  \includegraphics[width=\linewidth]{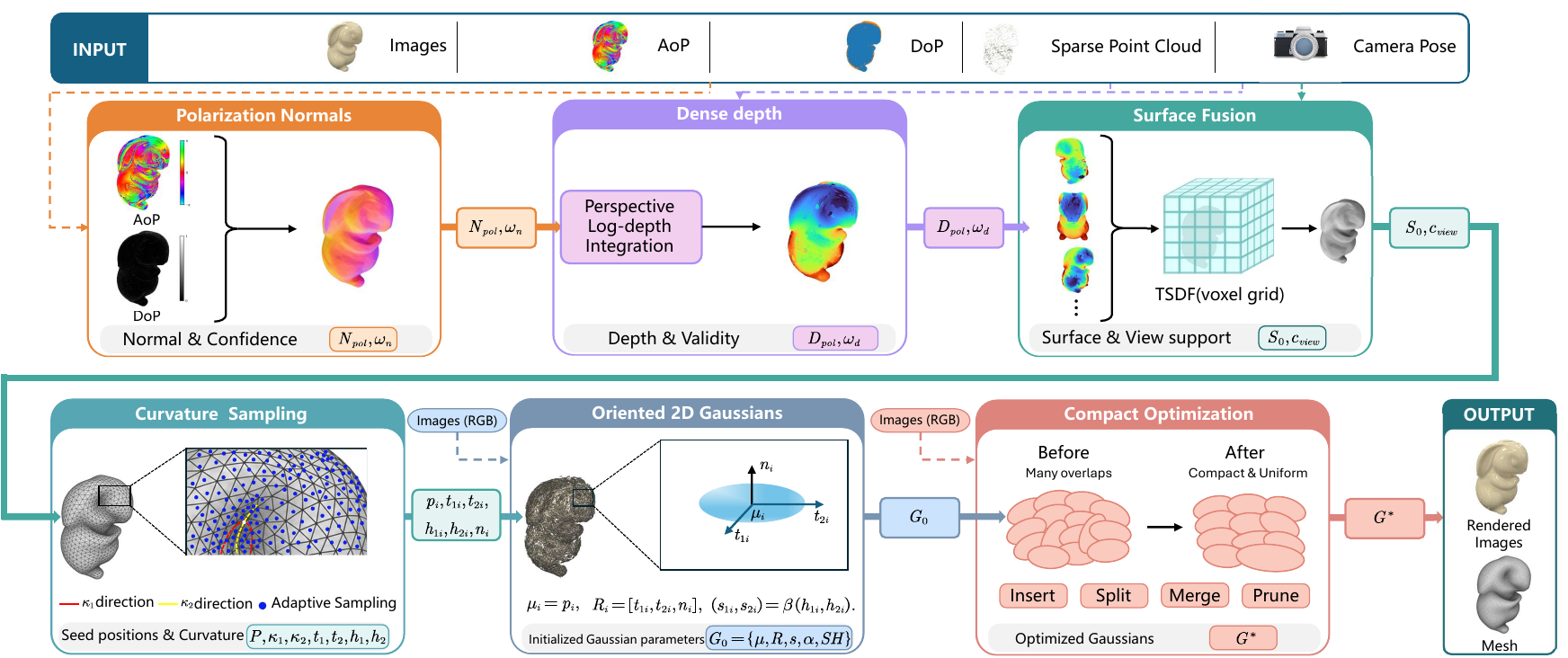}
  \par\vspace{-5pt} 
  \caption{\textbf{Overview.} Our method recovers surface normals from multi-view polarization observations, integrates them into dense depth maps using sparse depth anchors, and initializes oriented 2D Gaussians through TSDF surface fusion and curvature-adaptive sampling. Surface-constrained optimization and unified Insert, Split, Merge, and Prune updates then refine the representation for novel-view synthesis and mesh reconstruction.}
  \label{fig1}
  \par\vspace{-8pt} 
\end{figure*}

\subsection{Novel View Synthesis and Mesh Extraction}
Neural radiance fields (NeRF)~\cite{nerf2020} synthesize novel views by integrating volume density and view-dependent radiance along camera rays. However, a density field can reproduce images without identifying a unique surface~\cite{neus2021}. NeuS~\cite{neus2021} addresses this ambiguity by representing the surface as the zero level set of a learned signed distance function. 3D Gaussian splatting (3DGS)~\cite{kerbl2023gaussians} instead rasterizes anisotropic Gaussians for real-time rendering, but the primitives do not directly define a mesh.

Gaussian reconstruction methods differ in how they recover a surface from these primitives. SuGaR~\cite{sugar2024} aligns Gaussians with the surface and applies Poisson reconstruction~\cite{poisson2006} to sampled points. 2DGS~\cite{huang2024_2dgs} represents local surfaces as oriented disks and extracts a mesh by fusing rendered depth maps. Gaussian Opacity Fields (GOF)~\cite{gof2024} extracts an opacity isosurface on an adaptive tetrahedral grid. MILo~\cite{milo2025} incorporates differentiable mesh extraction into training, allowing discrepancies between mesh and Gaussian renderings to guide optimization. Sparse2DGS~\cite{sparse2dgs2025} exploits geometry at an earlier stage by initializing primitives from stereo depth maps and selectively updating them using rendered geometry. Inspired by this approach, we instead integrate polarization-derived normals and use curvature-adaptive sampling to initialize oriented primitives.

\subsection{Polarization-Guided 3D Reconstruction}
Polarization provides surface orientation cues in weakly textured regions, but orientation alone does not determine depth. The method in Ref.~\cite{cui2017polarization} employs polarization-derived azimuths to trace iso-depth contours and propagate reliable stereo depths. PANDORA~\cite{pandora2022} incorporates polarization into neural inverse rendering to recover shape and illumination while separating diffuse and specular radiance. NeRSP~\cite{nersp2024} combines supervision on rendered Stokes vectors with multi-view azimuth consistency for sparse-view reconstruction. PISR~\cite{pisr2024} directly constrains implicit surface normals using the polarization angle and the viewing ray at each pixel.

Gaussian methods exploit polarization at different stages. PolGS~\cite{polgs2025} uses measured Stokes vectors to supervise diffuse and specular reflection, while PhyGaP~\cite{phygap2026} uses polarimetric deferred rendering and self-occlusion-aware environment maps for reflection decomposition and relighting. PolarGS~\cite{polargs2026} incorporates polarization into PatchMatch~\cite{patchmatch2009} depth completion and back-projects the completed depths to add Gaussians. Our Pol2DGS uses polarization to establish surface coverage before Gaussian optimization. Sparse geometric anchors fix the depth scale during normal integration. On the fused surface, local normals and principal directions determine disk orientation, while principal curvatures guide sampling intervals and disk scales.
\section{Method}


\cref{fig1} presents the overview of Pol2DGS. The input, as illustrated, comprises multiview red-green-blue (RGB) images, angle of linear polarization (AoLP) and degree of linear polarization (DoLP) measurements, foreground and polarization validity masks, fixed camera calibration, and feature tracks.

\subsection{Polarization-Guided Surface Recovery}

\begin{figure}[t]
  \centering
  \includegraphics[width=\linewidth]{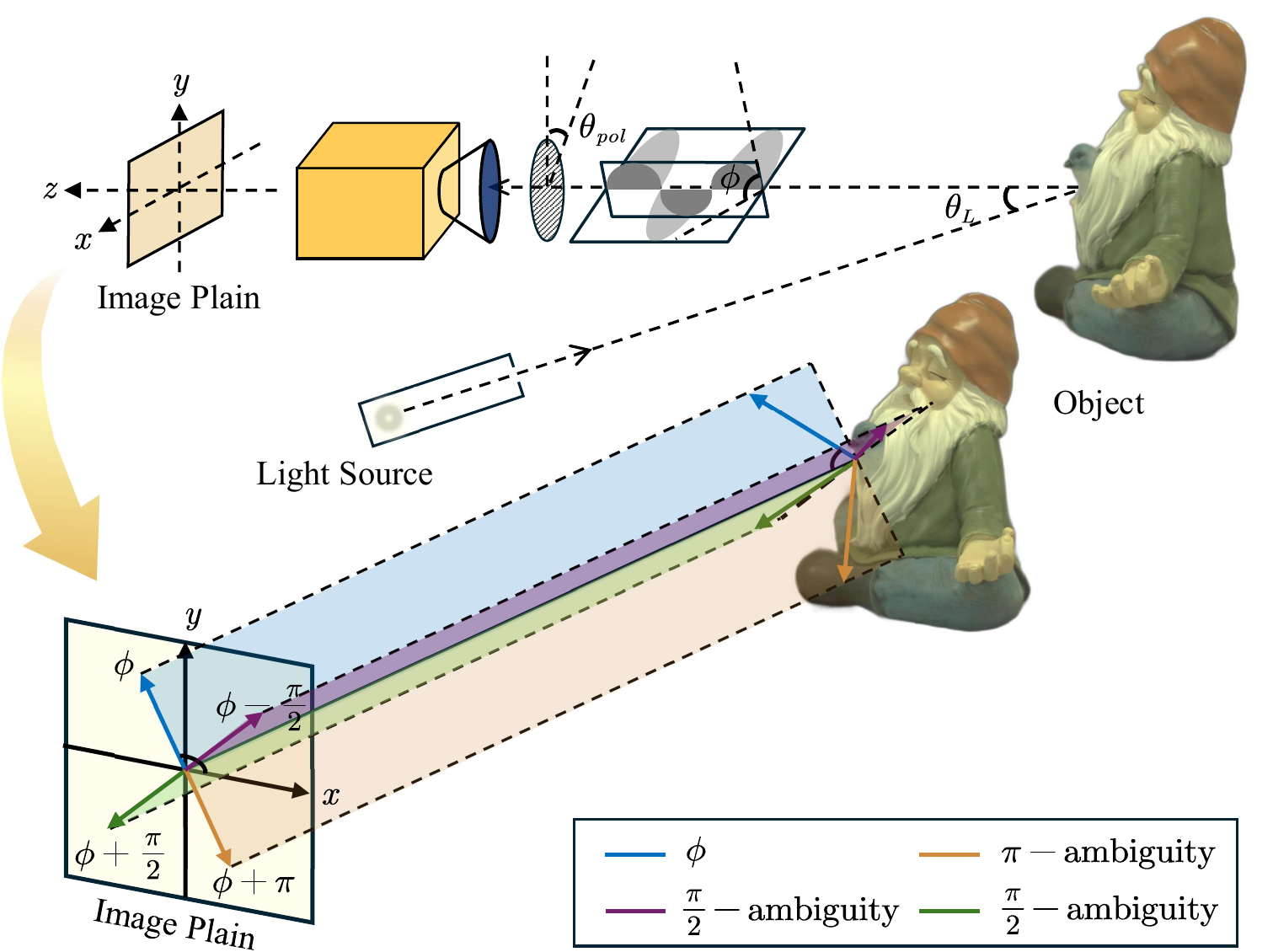}
  \par\vspace{-5pt} 
  \caption{\textbf{Four possible azimuth angles.} The figure illustrates the $\pi/2$ and $\pi$ ambiguities.}
  \label{fig2}
  \par\vspace{-8pt} 
\end{figure}

As shown in \cref{fig2}, polarization observations provide local directional cues about surfaces but exhibit two types of directional ambiguity. First, linear polarization defines an unoriented axis, making the observed angle of linear polarization (AoLP) $\phi$ indistinguishable from $\phi+\pi$. This gives rise to a $\pi$ ambiguity. Second, under standard reflection models, diffuse and specular reflection produce orthogonal polarization directions. An additional $\pi/2$ ambiguity therefore arises when the reflection mechanism is unknown~\cite{pmvir2023,miyazaki2012polarization}. Moreover, recovered normals cannot directly determine surface positions~\cite{normalintegration2018}.

For a valid observation, let $\mathbf{n}$ denote the unit surface normal in camera coordinates and $\hat{\mathbf{r}}$ the unit ray from the camera toward the surface. To relate the 3D normal to the polarization observation, we predict the polarization azimuth from the plane of incidence defined by these two vectors~\cite{pisr2024}:
\begin{equation}
\mathbf{m}=\mathbf{n}\times\hat{\mathbf{r}},
\qquad
\alpha=\operatorname{atan2}(-m_y,m_x).
\label{eq:polarization_azimuth}
\end{equation}
where $\mathbf{m}$ is the normal to the plane of incidence, $m_x,m_y$ are its two components, and $\alpha$ is the predicted polarization azimuth. The minus sign follows the coordinate conventions of the camera and AoLP images. Each surface normal is jointly constrained by its corresponding multiview observations.

To account for these ambiguities, we define a periodic angular difference and its penalty for the predicted azimuth $\alpha$ and observed AoLP $\phi$:
\begin{equation}
\begin{aligned}
h(\alpha,\phi)
&= \min_{j\in\mathbb{Z}}
   \left|\alpha-\phi-j\frac{\pi}{2}\right|, \\
C_k(\alpha,\phi)
&= \left[
   \frac{e^{-k(1-4h/\pi)}-e^{-k}}
        {1-e^{-k}}
   \right]^2.
\end{aligned}
\label{eq:polarization_angular_penalty}
\end{equation}
where the integer $j$ enumerates periodically equivalent azimuths, and $h\in[0,\pi/4]$ is the minimum angular difference. The parameter $k>0$ controls the penalty shape, and $C_k$ measures the inconsistency between the prediction and observation. We use the observation-weighted mean of $C_k$ as the polarization data term. An orientation constraint suppresses back-facing normals, while a smoothness term between neighboring vertex normals reduces unstable local variation. The degree of linear polarization (DoLP) is used to reject unreliable observations and assign weights~\cite{polarpms2024,pmvir2023,polarguide2026}. The normal variables shared across views and these constraints jointly reduce polarization direction ambiguities.

For each view, we obtain normalized image coordinates $(x,y)$ by subtracting the principal point $(c_x,c_y)$ from pixel coordinates $(u,v)$ and dividing by the corresponding focal lengths $f_x,f_y$. Each surface point lies along its viewing ray at depth $z$, measured along the camera’s optical axis, and we use its log depth $\ell$ for subsequent integration.

The orthogonality of surface normals and tangent vectors gives the following log-depth gradients under perspective projection~\cite{normalintegration2018}:
\begin{equation}
\begin{aligned}
g_u
&= \frac{\partial\ell}{\partial u}
 = -\frac{n_x}{f_x(n_x x+n_y y+n_z)}, \\
g_v
&= \frac{\partial\ell}{\partial v}
 = -\frac{n_y}{f_y(n_x x+n_y y+n_z)}.
\end{aligned}
\label{eq:log_depth_gradients}
\end{equation}
where $n_x,n_y,n_z$ are the components of the unit normal in this view, and $g_u,g_v$ describe log-depth changes along the two image coordinate directions. Pixels with near-zero denominators are excluded because the conversion is unstable there.

Matching these gradients alone cannot fix the depth of each independent region. We therefore introduce geometric anchors. Sparse anchors are obtained by retriangulating feature tracks from the training views and filtering them using reprojection and foreground support checks. For each independent region without sparse anchors, we add one reference depth anchor from the visual hull~\cite{visualhull1994} to fix its remaining depth degree of freedom. Depth is recovered for each view by solving
\begin{equation}
\begin{aligned}
\hat{\ell}
=
\arg\min_{\ell}
\sum_{(p,q)\in E}
\left(\ell_q-\ell_p-\Delta_{pq}\right)^2
+
\lambda_A
\sum_{a\in A}
\left(\ell_a-\log Z_a\right)^2.
\end{aligned}
\label{eq:anchored_depth_integration}
\end{equation}

\noindent where $p,q$ are adjacent valid pixels, and $E$ is the set of valid adjacency edges. The predicted log-depth difference $\Delta_{pq}$ is computed from the average gradient at the endpoints and the actual pixel spacing. The set $A$ contains depth anchors, and $Z_a$ is the reference depth associated with anchor index $a$. The parameter $\lambda_A$ controls anchoring strength, and $\hat{\ell}$ is the optimal log depth.

To avoid propagating gradients across surface discontinuities, we remove adjacency edges where the visual-hull reference depth changes abruptly. We solve this least-squares problem with LSQR~\cite{lsqr1982} and then exponentiate the solution to recover depth.

Finally, we fuse the per-view depths with equal weights into a TSDF~\cite{curless1996volumetric}, excluding observations outside the foreground and those with invalid depth. Multiview support checks and connected-component processing then produce a unified surface. 

\subsection{Surface-Adaptive Initialization of Oriented Gaussians}

\begin{table*}[t]
    \centering
    \caption{
        \textbf{Quantitative results of different methods on mesh reconstruction.}
        Red, orange, and yellow indicate the best, second-best, and third-best results, respectively.
    }
    \label{tab1}
    \par\vspace{-5pt} 

    \begingroup
    \fontsize{10pt}{12pt}\selectfont

    \setlength{\tabcolsep}{2pt}
    \renewcommand{\arraystretch}{1.05}

    \setlength{\aboverulesep}{1.5pt}
    \setlength{\belowrulesep}{1.5pt}

    \renewcommand{\tabularxcolumn}[1]{m{#1}}

    \begin{tabularx}{\textwidth}{
        @{}cl*{8}{>{\centering\arraybackslash}X}c
    }
        \toprule
        & \multirow{2}{*}{Method}
        & \multicolumn{2}{c}{B.\ Loong}
        & \multicolumn{2}{c}{R.\ Loong}
        & \multicolumn{2}{c}{L.\ Rabbit}
        & \multicolumn{2}{c}{S.\ Rabbit}
        & \multirow{2}{*}{Time} \\
        &
        & CD $\downarrow$ & \mbox{F-Score $\uparrow$}
        & CD $\downarrow$ & \mbox{F-Score $\uparrow$}
        & CD $\downarrow$ & \mbox{F-Score $\uparrow$}
        & CD $\downarrow$ & \mbox{F-Score $\uparrow$} \\
       
        \cmidrule(r{4pt}){1-2}
        \cmidrule(l{4pt}r{4pt}){3-10}
        \cmidrule(l{4pt}){11-11}

        \multirow{3}{*}{%
            \rotatebox{90}{%
                \scriptsize
                \shortstack{Polarimetric\\Implicit}%
            }%
        }
        & PANDORA~\cite{pandora2022}
        & 2.49 & 39.34\%
        & 2.87 & 44.25\%
        & 1.50 & 55.27\%
        & 1.53 & \thirdbest{60.79\%}
        & 12h+ \\

        & NeRSP~\cite{nersp2024}
        & 2.42 & 41.35\%
        & 2.38 & \thirdbest{47.74\%}
        & \best{1.34} & \secondbest{68.66\%}
        & \best{0.86} & \best{77.69\%}
        & 12h+ \\

        & PISR~\cite{pisr2024}
        & \best{1.19} & \best{54.50\%}
        & \best{1.64} & \best{72.88\%}
        & \secondbest{1.36} & \best{75.92\%}
        & \secondbest{0.89} & \secondbest{75.55\%}
        & \secondbest{24.5m} \\

        \cmidrule(r{4pt}){1-2}
        \cmidrule(l{4pt}r{4pt}){3-10}
        \cmidrule(l{4pt}){11-11}

        \multirow{7}{*}{\rotatebox{90}{Explicit}}
        & SuGaR~\cite{sugar2024}
        & \thirdbest{2.01} & \thirdbest{34.85\%}
        & 2.83 & 28.65\%
        & 1.46 & 46.55\%
        & 1.78 & 41.42\%
        & 43.0m \\

        & 2DGS~\cite{huang2024_2dgs}
        & 2.46 & 26.54\%
        & 2.41 & 23.70\%
        & \thirdbest{1.43} & 40.43\%
        & 2.41 & 34.03\% 
        & 25.2m \\

        & GOF~\cite{gof2024}
        & 4.10 & 18.21\%
        & 4.15 & 19.52\%
        & 2.93 & 27.90\%
        & 2.91 & 30.03\% 
        & 27.2m \\

        & MILo~\cite{milo2025}
        & 4.08 & 15.13\%
        & 2.19 & 28.17\%
        & 1.52 & 40.92\%
        & 1.76 & 34.34\% 
        & 33.1m \\

        & Sparse2DGS~\cite{sparse2dgs2025}
        & 8.99 & 21.20\%
        & 7.83 & 18.62\%
        & 7.32 & 22.63\%
        & 7.43 & 22.21\% 
        & 6.1m{%
          \renewcommand{\thefootnote}{\fnsymbol{footnote}}%
          \footnotemark[1]%
        } \\

        & PhyGaP~\cite{phygap2026}
        & 2.22 & 33.69\%
        & \thirdbest{2.15} & 40.74\%
        & 1.53 & 55.81\%
        & \thirdbest{1.41} & 58.22\%
        & \thirdbest{21.3m} \\

        & \textbf{Pol2DGS (Ours)}
        & \secondbest{1.95} & \secondbest{40.46\%}
        & \secondbest{1.95} & \secondbest{48.06\%}
        & \thirdbest{1.43} & \thirdbest{56.86\%}
        & 1.66 & 49.28\% 
        & \best{4.3m} \\

        \bottomrule
    \end{tabularx}
    \endgroup
    \par\vspace{-8pt} 
\end{table*}

\begin{figure*}[t]
  \centering
  \includegraphics[width=\linewidth]{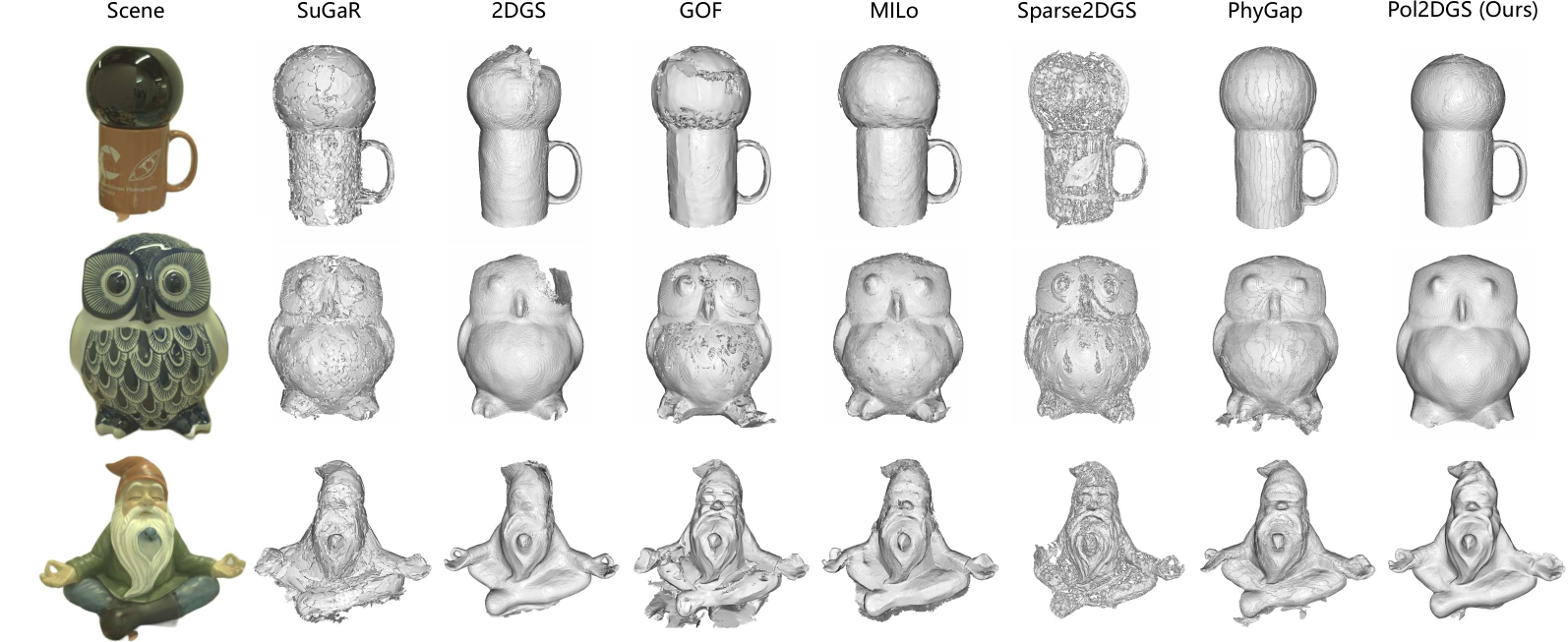}
  \par\vspace{-5pt} 
  \caption{\textbf{Mesh reconstruction results of different methods.} Meshes extracted from three real-world scenes in the PANDORA~\cite{pandora2022} dataset.}
  \label{fig3}
  \par\vspace{-8pt} 
\end{figure*}
Primitives can cover larger areas in relatively flat regions, while curved regions require finer resolution. The two sides of an edge may also have different surface orientations. To accommodate these differences, we take the fused surface and its support information as input. We determine sampling positions from local curvature and edge structure, then use these samples to initialize the orientations and scales of 2D Gaussians.

In local tangent coordinates $(\xi,\eta)$, we fit the normal height $\zeta$ using a local polynomial approximation~\cite{cazals2005}:

\begin{equation}
\begin{gathered}
\zeta(\xi,\eta)=\tfrac12 a\xi^2+b\xi\eta+\tfrac12 c\eta^2+d\xi+e\eta+f,\\
H=\begin{bmatrix}a&b\\b&c\end{bmatrix}.
\end{gathered}
\label{eq:local_quadratic_surface}
\end{equation}

{%
  \renewcommand{\thefootnote}{\fnsymbol{footnote}}%
  \footnotetext[1]{Sparse2DGS did not produce complete object
  reconstructions in our tests. We therefore exclude it
  from the runtime ranking.}%
}

\indent The fit uses distance and support weights with one robust reweighting step.
Under a local small-slope approximation, the eigenvalues of $H$ approximate principal curvatures $\kappa_1,\kappa_2$, and its eigenvectors define principal directions.

The directional sampling spacings are
\begin{equation}
h_j=\operatorname{clip}\!\left(
\sqrt{\frac{8\epsilon}{|\kappa_j|+\delta}}\frac{1}{\sqrt{1+(\theta/\theta_0)^2}},
h_{\min},h_{\max}\right),
\label{eq:adaptive_sampling_spacing}
\end{equation}
where $j\in\{1,2\}$, $\epsilon$ is the height-error tolerance, $\delta>0$ stabilizes the denominator, and $\theta$ is the root-mean-square angle of neighborhood normal variation.
For fusion voxel size $v$, we set $\epsilon=0.5v$, $h_{\min}=v$, $h_{\max}=5v$, and $\theta_0=15^\circ$.
In regular regions, candidates are allocated by triangle area and local spacing, followed by variable-radius Poisson-disk rejection~\cite{corsini2012} with scalar exclusion radius
\begin{equation}
r=\sqrt{h_1h_2}.
\label{eq:sampling_exclusion_radius}
\end{equation}
where $r$ is a scalar exclusion radius, and $h_1,h_2$ are the sampling spacings along the two tangential directions. Point exclusion is based on $r$, while the directional scale differences are retained for subsequent Gaussian initialization.

We detect sharp edges by adjacent-face dihedral angles and partition the surface while accounting for boundary and nonmanifold edges. Samples on the two sides of each sharp edge are generated separately using their respective normals.

After transforming samples and their directions and spacings into training coordinates, we initialize
\begin{equation}
\boldsymbol{\mu}=\mathbf p,\quad
R=[\mathbf t_1,\mathbf t_2,\mathbf n],\quad
(s_1,s_2)=\gamma(h_1,h_2),
\label{eq:gaussian_initialization}
\end{equation}
where $\mathbf p$ is the sample position and $(\mathbf t_1,\mathbf t_2,\mathbf n)$ is an orthonormal frame with surface normal $\mathbf n$.
Tangents follow principal directions in regular regions; edge samples use tangent planes defined by their respective side normals.
We set $\gamma=0.8$ and opacity to $0.1$, initialize zeroth-order spherical harmonics~\cite{plenoctrees2021,plenoxels2022,kerbl2023gaussians} from fused surface colors, and set higher orders to zero.
Sample positions, reference normals, and support confidences serve as surface anchors for subsequent optimization; initial rotations and scales define reference ranges for geometry updates.

\subsection{Geometry-Guided Primitive Adaptation}

We first freeze positions, rotations, and scales, optimizing only spherical harmonic coefficients and opacity. We then enable geometry updates with a lower learning rate. After each update, center displacement, relative rotation, and scale changes are bounded by their respective allowed ranges to control deviations from the reference surface. We also remove the original Clone operation~\cite{kerbl2023gaussians,huang2024_2dgs} and manage the primitive set through Insert, Split, Merge, and Prune.

\textbf{Insert} targets regions with persistently low coverage. We recover patches through robust integration using polarization-derived normals and reliable boundary depths. After confirming support from at least two other views and the absence of conflicts, we sample these patches to generate new Gaussians. We first check their projected extents, then insert them temporarily and render the scene. The new primitives are retained if coverage increases in missing regions and background coverage, image error, and depth changes remain within their allowed limits. Otherwise, they are removed. We establish surface anchors from their patches, temporarily freeze their geometry, and apply a protection period for primitive updates.

\textbf{Split} refines the local representation provided by existing primitives. Candidates must have large average screen-space gradients over recent training iterations. They must also exhibit at least one of the following conditions: excessive projected coverage, substantial scale growth relative to initialization, large neighborhood normal variation, or a large deviation from the reference tangent plane. Observation-count and opacity criteria must also be satisfied. We construct two child Gaussians along the major axis in the tangent plane:
\begin{equation}
\begin{gathered}
\boldsymbol{\mu}_{\pm}
= \boldsymbol{\mu}
  \pm 0.45s_{\max}\mathbf{t}_{\max}, \\
s_{\max,\pm}=0.65s_{\max},
\qquad
o_{\pm}=1-\sqrt{1-o}.
\end{gathered}
\label{eq:gaussian_split}
\end{equation}
where $\boldsymbol{\mu}$ and $o$ are the parent's center and opacity, respectively. The quantity $s_{\max}$ is the larger of the two tangential scales, and $\mathbf{t}_{\max}$ is the corresponding unit tangent direction. The subscripts $+$ and $-$ identify the two children. The children inherit the parent's minor-axis scale and orientation and replace the parent. The opacity adjustment reduces changes in local coverage caused by splitting.

\textbf{Merge} combines redundant primitives that are spatially close, have aligned normal axes, and are approximately coplanar. We use opacity and coverage area to fit a new center and scales. The merged primitive retains the orientation and anchor of the parent with higher opacity. Its opacity is fitted to the combined coverage of both parents. To measure the local coverage loss caused by merging, we define
\begin{equation}
\begin{gathered}
A_{\mathrm{old}}(x)
= a_1(x)+a_2(x)-a_1(x)a_2(x), \\
E_{\mathrm{cov}}
= \frac{
    \sum_x
    \max\left\{
        A_{\mathrm{old}}(x)-A_{\mathrm{new}}(x),\,0
    \right\}
}{
    \sum_x A_{\mathrm{old}}(x)
}.
\end{gathered}
\label{eq:merge_coverage_loss}
\end{equation}
where $x$ is a sample location in the local tangent plane, and $a_1,a_2$ are the opacity contributions of the two parents. The quantities $A_{\mathrm{old}}$ and $A_{\mathrm{new}}$ represent coverage before and after merging, respectively, and $E_{\mathrm{cov}}$ measures the relative loss of local coverage. A merge is accepted if this ratio does not exceed $\tau_{\mathrm{cov}}$ and the geometric changes remain within their allowed ranges. Each primitive can participate in at most one merge pair per update.

\textbf{Prune} removes primitives whose opacity falls below a threshold while retaining at least one primitive.\\
\indent The four operations jointly maintain a single Gaussian set while synchronously updating its parameters, surface anchors, and optimizer states. Meshes are extracted by fusing depth maps rendered from the optimized Gaussian model~\cite{huang2024_2dgs,curless1996volumetric}.

\subsection{Training Loss}

The total training loss comprises five terms:
\begin{equation}
\mathcal{L}
=
\mathcal{L}_{\mathrm{rgb}}
+\lambda_1\mathcal{L}_{\mathrm{normal}}
+\lambda_2\mathcal{L}_{\mathrm{p2pl}}
+\lambda_3\mathcal{L}_{\mathrm{mask}}
+\lambda_4\mathcal{L}_{\mathrm{smooth}}.
\label{eq:training_loss}
\end{equation}
where $\mathcal{L}_{\mathrm{rgb}}$ is the foreground image reconstruction loss, $\mathcal{L}_{\mathrm{normal}}$ enforces consistency between rendered normals and depth-derived normals~\cite{huang2024_2dgs}, and $\mathcal{L}_{\mathrm{p2pl}}$ constrains the distance from Gaussian centers to their reference tangent planes. The mask loss $\mathcal{L}_{\mathrm{mask}}$ suppresses opacity coverage outside the foreground silhouette, while $\mathcal{L}_{\mathrm{smooth}}$ includes normal and depth smoothness constraints. The weights $\lambda_1,\ldots,\lambda_4$ control the corresponding loss terms.

\section{Experiments}
\label{sec:Experiments}

\subsection{Implementation and Dataset}
\label{sec:pol2dgs-implementation}
We train for 15000 iterations, freezing geometry for the first 1500. We set $\lambda_1=0.5$, $\lambda_2=0.3$, $\lambda_3=0.1$, and $\lambda_4=1$; $\mathrm{L}_{smooth}$ combines normal and depth terms with weights 0.2 and 0.8, respectively. We select seven scenes from two datasets: vase, owl, and gnome from PANDORA~\cite{pandora2022}, and BlackLoong, RedLoong, LayingRabbit, and StandingRabbit from PISR~\cite{pisr2024}. Both datasets contain real-world captures, and PISR additionally provides ground-truth meshes obtained using structured-light scanning.
\subsection{Comparison}
\label{sec:pol2dgs-comparison}

\subsubsection{Mesh Reconstruction}

\begin{figure*}[t]
  \centering
  \includegraphics[width=\linewidth]{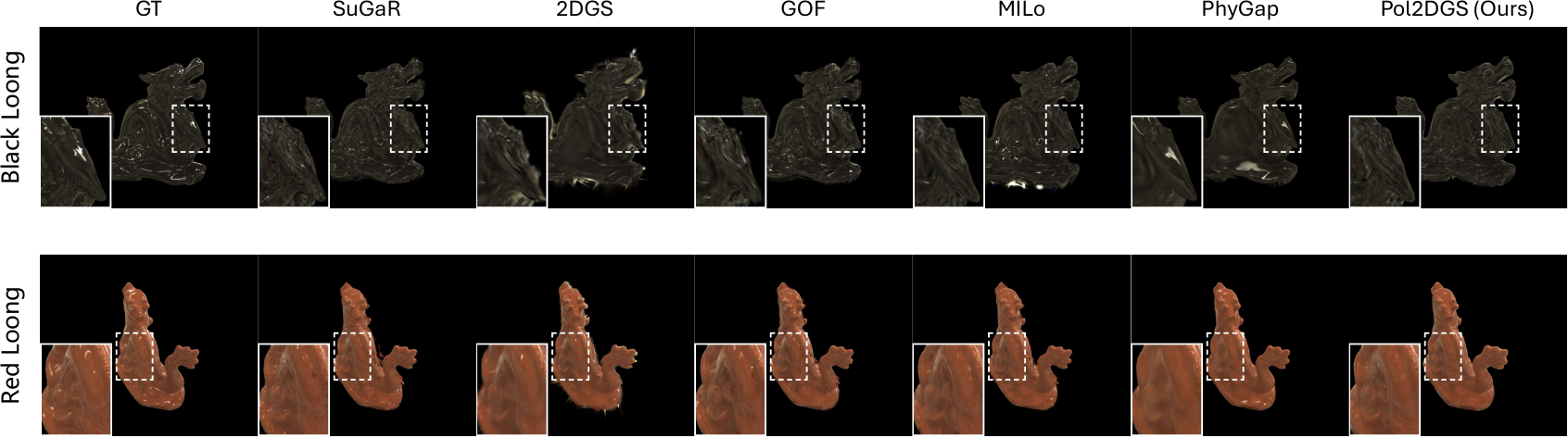}
  \par\vspace{-5pt} 
  \caption{\textbf{Qualitative results of different methods on novel view synthesis.} We zoom in on BlackLoong's forelimb and RedLoong's body to assess geometric reconstruction accuracy in regions with complex details.}
  \label{fig4}
  \par\vspace{-8pt} 
\end{figure*}

As shown in \Cref{tab1}, Pol2DGS achieves the shortest reported runtime (4.3min), yielding a 5× speedup over PhyGaP~\cite{phygap2026} while improving both CD and F-score on three of the four scenes. Compared with polarization-guided implicit methods~\cite{pandora2022,nersp2024,pisr2024}, our method substantially reduces training time while maintaining competitive reconstruction quality. It is important to appreciate that, Pol2DGS further outperforms PhyGaP on Black Loong and Red Loong, which contain complex surface structures and fine-scale geometric details. These gains are consistent with the role of polarization-derived surface priors in guiding Gaussian placement and orientation before optimization. Curvature-adaptive sampling allocates samples according to local geometric variation, providing finer coverage in geometrically complex regions~\cite{pauly2002simplification,huang2013ear,huang2009consolidation}. Further strong evidence that our design may help preserve fine surface structures within a limited optimization budget is shown in \cref{fig3}.  As shown, Pol2DGS avoids overfitting surface patterns as geometry and produces cleaner, more coherent surfaces than the other competing methods. The overall shape and key geometric structures are likewise preserved.

\subsubsection{Novel View Synthesis}
\cref{fig4} presents a comparison of novel-view renderings across the evaluated methods. The competing Gaussian-based approaches exhibit varying degrees of rendering artifacts, while PhyGaP~\cite{phygap2026} produces overly smooth results in some regions. Pol2DGS better preserves fine appearance details and sharp image boundaries in the illustrated examples. As shown in \Cref{tab2}, Pol2DGS achieves the best or second-best scores in most scene-specific PSNR and SSIM~\cite{ssim2004} comparisons, demonstrating its competitive image reconstruction quality.

The performance differences can be attributed to the fact that, PhyGaP~\cite{phygap2026} enables relighting through polarimetric deferred rendering and self-occlusion-aware environment map construction. While these components improve novel-view synthesis quality, they do not substantially improve mesh reconstruction quality and incur additional computational overhead. Pol2DGS focuses on using surface priors to guide Gaussian initialization and optimization. Beyond the numbers reported and visual comparisons, it is a delicate balance between geometric accuracy and optimization efficiency.


\begin{table}[t]
    \centering
    \caption{
        \textbf{Quantitative results of different methods on novel view synthesis.} 
    }
    \label{tab2}
    \par\vspace{-5pt} 

    \begingroup
    \fontsize{8pt}{9.5pt}\selectfont
    \setlength{\tabcolsep}{0.6pt}
    \renewcommand{\arraystretch}{1.08}

    \setlength{\aboverulesep}{1pt}
    \setlength{\belowrulesep}{1pt}

    \renewcommand{\tabularxcolumn}[1]{m{#1}}

    \begin{tabularx}{\columnwidth}{
        @{}cc*{6}{>{\centering\arraybackslash}X}
    }
        \toprule
        \textbf{Scene}
        & \textbf{Metric}
        & {\scriptsize\shortstack{SuGaR\\\cite{sugar2024}}}
        & {\scriptsize\shortstack{2DGS\\\cite{huang2024_2dgs}}}
        & {\scriptsize\shortstack{GOF\\\cite{gof2024}}}
        & {\scriptsize\shortstack{MILo\\\cite{milo2025}}}
        & {\scriptsize\shortstack{PhyGaP\\\cite{phygap2026}}}
        & {\scriptsize
            \shortstack{\textbf{Pol2DGS}\\\textbf{(Ours)}}
          } \\
        \midrule

        \multirow{2}{*}{B.\ Loong}
        & PSNR $\uparrow$
        & 19.78 & \thirdbest{20.03} & 19.82 & 17.56 & \best{21.04} & \secondbest{20.50} \\
        & SSIM $\uparrow$
        & 0.508 & \thirdbest{0.628} & 0.573 & 0.441 & \best{0.702} & \secondbest{0.635} \\

        \midrule[0.3pt]

        \multirow{2}{*}{R.\ Loong}
        & PSNR $\uparrow$
        & 21.68 & \thirdbest{22.75} & 22.15 & 18.25 & \best{23.76} & \secondbest{23.59} \\
        & SSIM $\uparrow$
        & 0.655 & \thirdbest{0.760} & 0.708 & 0.625 & \best{0.811} & \secondbest{0.775} \\

        \midrule[0.3pt]

        \multirow{2}{*}{L.\ Rabbit}
        & PSNR $\uparrow$
        & 17.79 & \thirdbest{24.30} & 22.71 & 16.12 & \secondbest{26.36} & \best{26.45} \\
        & SSIM $\uparrow$
        & 0.759 & \thirdbest{0.882} & 0.841 & 0.717 & \best{0.920} & \secondbest{0.873} \\

        \midrule[0.3pt]

        \multirow{2}{*}{S.\ Rabbit}
        & PSNR $\uparrow$
        & 11.39 & \thirdbest{21.80} & 21.36 & 15.10 & \best{25.67} & \secondbest{25.43} \\
        & SSIM $\uparrow$
        & 0.536 & \thirdbest{0.857} & 0.810 & 0.658 & \best{0.903} & \secondbest{0.849} \\

        \bottomrule
    \end{tabularx}
    \endgroup
    \par\vspace{-8pt} 
\end{table}


\subsection{Ablations}

\begin{table}[t]
    \centering
    \caption{
        \textbf{Ablation study of Pol2DGS on BlackLoong.} Polarimetric surface priors improve both geometric accuracy and rendering quality over 2DGS~\cite{huang2024_2dgs}. 
    }
    \label{tab3}
    \par\vspace{-5pt} 

    \begingroup
    \fontsize{8pt}{9.5pt}\selectfont
    \setlength{\tabcolsep}{1.2pt}
    \renewcommand{\arraystretch}{1.1}

    \setlength{\aboverulesep}{1pt}
    \setlength{\belowrulesep}{1pt}

    \begin{tabular*}{\columnwidth}{
        @{\extracolsep{\fill}}ccccccccc@{}
    }
        \toprule
        & 2DGS~\cite{huang2024_2dgs}
        & \shortstack{Polarimetric\\Surface Priors}
        & Insert
        & Split
        & Merge
        & Prune
        & CD $\downarrow$
        & PSNR $\uparrow$ \\
        \midrule

        \textbf{(a)}
        & $\checkmark$ & & & & &
        & 2.46 & 20.03 \\

        \textbf{(b)}
        & $\checkmark$ & $\checkmark$ & & & &
        & 2.09 & \textbf{20.51} \\

        \textbf{(c)}
        & $\checkmark$ & $\checkmark$ & & $\checkmark$ & $\checkmark$ & $\checkmark$
        & 2.01 & 20.37 \\

        \textbf{(d)}
        & $\checkmark$ & $\checkmark$ & $\checkmark$ & & $\checkmark$ & $\checkmark$
        & 2.14 & 20.48 \\

        \textbf{(e)}
        & $\checkmark$ & $\checkmark$ & $\checkmark$ & $\checkmark$ & & $\checkmark$
        & 2.00 & 20.37 \\

        \textbf{(f)}
        & $\checkmark$ & $\checkmark$ & $\checkmark$ & $\checkmark$ & $\checkmark$ &
        & 2.10 & 20.47 \\

        \textbf{(g)}
        & $\checkmark$ & $\checkmark$ & $\checkmark$ & $\checkmark$ & $\checkmark$ & $\checkmark$
        & \textbf{1.97} & 20.50 \\

        \bottomrule
    \end{tabular*}
    \endgroup
    \par\vspace{-8pt} 
\end{table}

We conduct ablation studies to evaluate the contributions of polarization-derived geometric priors and geometry-guided primitive adaptation. \Cref{tab3} shows that incorporating geometric priors substantially improves overall reconstruction performance, providing an effective starting point for Gaussian optimization. Building on this initialization, the primitive adaptation strategy further improves geometric accuracy and novel-view synthesis quality by adjusting the primitive distribution to local surface requirements. These findings highlight the complementary roles of the two components. The geometric priors guide the initial positions, orientations, and scales of the Gaussians, while the adaptation strategy refines their distribution during optimization to better represent surface structure and appearance.

\begin{table}[t]
    \centering
    \caption{\textbf{Effect of polarization-derived normal supervision on normal accuracy.} We report the mean angular error (MAE), which is lowest across all four scenes without the polarization-derived normal constraint.}
    \label{tab4}
    \par\vspace{-5pt} 

    \begingroup
    \fontsize{8pt}{9.5pt}\selectfont
    \setlength{\tabcolsep}{0.8pt}
    \renewcommand{\arraystretch}{1.15}

    \setlength{\aboverulesep}{1pt}
    \setlength{\belowrulesep}{1pt}

    \begin{tabularx}{\columnwidth}{
        @{}l*{4}{>{\centering\arraybackslash}X}
    }
        \toprule
          & B.Loong
          & R.Loong
          & L.Rabbit
          & S.Rabbit \\
        \midrule

        w/ Pol. Normal Constraint & $27.351^\circ$ & $22.925^\circ$ & $19.638^\circ$ & $22.368^\circ$ \\
        w/o Pol. Normal Constraint & $\boldsymbol{23.257^\circ}$ & $\boldsymbol{21.976^\circ}$ & $\boldsymbol{16.129^\circ}$ & $\boldsymbol{19.824^\circ}$ \\
        Direct Depth Fusion & $26.608^\circ$ & $25.376^\circ$ & $21.833^\circ$ & $28.066^\circ$ \\

        \bottomrule
    \end{tabularx}
    \endgroup
    \par\vspace{-8pt} 
\end{table}

We further examine whether polarization-derived normals should continue to supervise Gaussian optimization. We compare three experimental settings: \textbf{(i)} Gaussian optimization is performed with an additional polarization-derived normal constraint. \textbf{(ii)} Gaussian optimization is performed without this constraint, using the same initialization and retaining all other optimization terms. \textbf{(iii)} Depth maps obtained by integrating the estimated normals are directly fused without subsequent Gaussian optimization. Experimentally, it finds that removing the additional constraint yields more accurate surface normals in the evaluated scenes. Referring to \Cref{tab4}, this clearly shows that polarization-derived normals provide useful geometric initialization but need not remain supervision targets throughout training. Subsequent refinement under photometric supervision and the geometric regularization can correct residual errors in the initialized surface.

\begin{figure}[t]
  \centering
  \includegraphics[width=\linewidth]{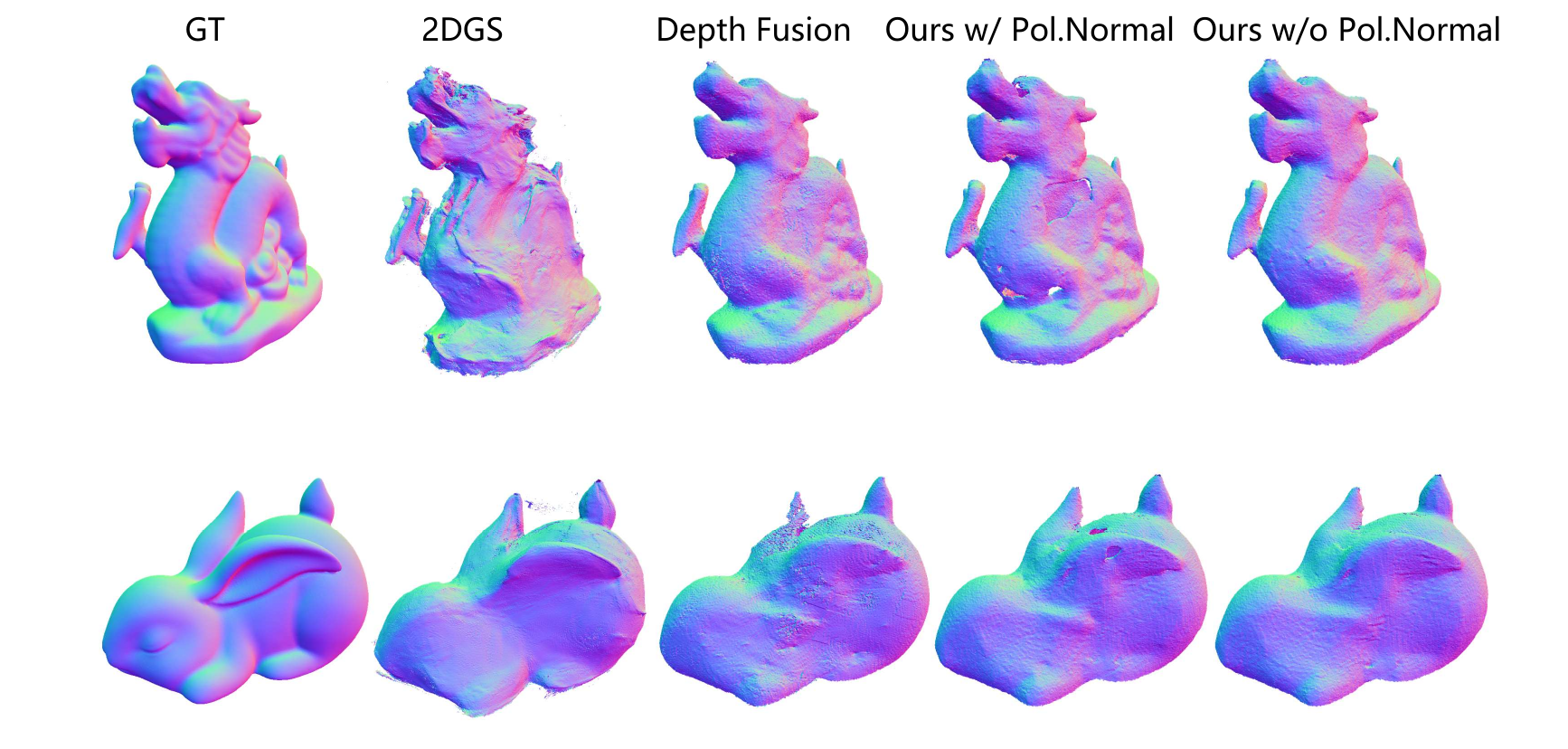}
  \par\vspace{-5pt} 
  \caption{\textbf{Qualitative results of normal maps under different supervision settings.} The normal map obtained without an additional polarization-derived normal constraint more closely matches the ground truth.}
  \label{fig5}
  \par\vspace{-8pt} 
\end{figure}

\begin{table}[t]
    \centering
    \caption{\textbf{Comparison of different initialization strategies.} Sampled polarization-based initialization achieves the lowest CD and shortest runtime across all four scenes.}
    \label{tab5}
    \par\vspace{-5pt} 

    \begingroup
    \fontsize{8pt}{9.5pt}\selectfont
    \setlength{\tabcolsep}{1pt}
    \renewcommand{\arraystretch}{1.1}

    \setlength{\aboverulesep}{1pt}
    \setlength{\belowrulesep}{1pt}

    \renewcommand{\tabularxcolumn}[1]{m{#1}}

    \begin{tabularx}{\columnwidth}{
        @{}lc*{3}{>{\centering\arraybackslash}X}
    }
        \toprule
        \textbf{Scene}
        &
        & \shortstack{Dense\\Pol.\ Init.}
        & \shortstack{Sparse COLMAP\\Init.~\cite{colmap2016}}
        & \shortstack{Sampled\\Pol.\ Init.} \\
        \midrule

        \multirow{2}{*}{B.\ Loong}
        & CD $\downarrow$
        & 2.04 & 2.19 & \textbf{1.95} \\
        & Time $\downarrow$
        & 349s & 291s & \textbf{267s} \\

        \addlinespace[2pt]

        \multirow{2}{*}{R.\ Loong}
        & CD $\downarrow$
        & 2.06 & 2.29 & \textbf{1.94} \\
        & Time $\downarrow$
        & 331s & 277s & \textbf{269s} \\

        \addlinespace[2pt]

        \multirow{2}{*}{L.\ Rabbit}
        & CD $\downarrow$
        & 1.61 & 1.91 & \textbf{1.44} \\
        & Time $\downarrow$
        & 338s & 277s & \textbf{253s} \\

        \addlinespace[2pt]

        \multirow{2}{*}{S.\ Rabbit}
        & CD $\downarrow$
        & 1.67 & 2.19 & \textbf{1.66} \\
        & Time $\downarrow$
        & 330s & 282s & \textbf{257s} \\

        \bottomrule
    \end{tabularx}
    \endgroup
    \par\vspace{-8pt} 
\end{table}

As a more fine-grained ablation, we compare, on the same evaluated scenes, three initialization from \textbf{(i)} dense depth maps integrated from polarization-derived normals, \textbf{(ii)} sparse COLMAP~\cite{colmap2016} points, and \textbf{(iii)} adaptively sampled surface points. As in \Cref{tab5}, Dense initialization improves geometric accuracy over COLMAP but increases optimization time. Adaptive sampling achieves higher accuracy and shorter optimization time than the dense alternative. Similar to the earlier finding that no additional constraint is needed for surface normal recovery, increasing the number of initial samples does not necessarily improve reconstruction quality and can introduce additional computational overhead. Selecting representative samples according to local surface geometry provides a more effective starting point for accurate and efficient reconstruction.

\section{Conclusion}
\label{sec:conclusion}
Pol2DGS accelerates 2D Gaussian surface reconstruction through geometric priors from polarization and primitive updates guided by geometry. Polarimetric normals, sparse geometric anchors, and anisotropic sampling adapted to local curvature initialize Gaussian positions, orientations, and scales without depth sensors. Experimentally, this initialization improves geometric accuracy and reduces optimization time. Insertion guided by surface geometry, deterministic splitting along the major axis, merging under coverage constraints, and pruning based on opacity further refine the representation. Ablations demonstrate that these operations jointly improve surface accuracy while largely preserving image quality. Across the evaluated scenes and under our timing protocol, optimization averages 4 minutes and 22 seconds. This yields a speedup of approximately 5 to 10 times over the mainstream methods while maintaining competitive mesh accuracy and novel view synthesis quality.
\section{LLM Usage Statement}
\label{sec:LLM}
Large Language Models (LLMs) were used to aid in the writing and polishing of the manuscript. Specifically, we used an LLM to assist in refining the language, improving readability, and ensuring clarity in various sections of the paper. The model helped with tasks such as sentence rephrasing, grammar checking, and enhancing the overall flow of the text. We also used the LLM to assist in identifying code vulnerabilities, improving code style, and ensuring the stability of program execution. The model helped with tasks such as logic correction, bug fixing, and optimizing the overall execution performance of the code.

It is important to note that the LLM was not involved in the ideation, research methodology, or experimental design. All research concepts, ideas, and analyses were developed and conducted by the authors. The contributions of the LLM were solely focused on improving the linguistic quality of the paper, with no involvement in the scientific content or data analysis.
{
    \small
    \bibliographystyle{ieeenat_fullname}
    \bibliography{main}
}


\end{document}